# Reference Resolution within the Framework of Cognitive Grammar

**Susanne Salmon-Alt**

*Laboratoire Loria, France*

**Laurent Romary**

*Laboratoire Loria, France*

Following the principles of Cognitive Grammar, we concentrate on a model for reference resolution that attempts to overcome the difficulties of previous approaches, based on the fundamental assumption that all reference (independent on the type of the referring expression) is accomplished via access to and restructuring of domains of reference rather than by direct linkage to the entities themselves. The model accounts for entities not explicitly mentioned but understood in a discourse, and enables exploitation of discursive and perceptual context to limit the set of potential referents for a given referring expression. As the most important feature, we note that our model can with a single mechanism handle what are typically treated as diverse phenomena. Our approach, then, provides a fresh perspective on the relations between Cognitive Grammar and the problem of reference.

## 1. Introduction

The work presented in this paper can be situated within a wider attempt to define man-machine dialogue systems that respects the fundamental communication means of their users. In particular, even when the underlying task to be achieved through such dialogue systems is limited (e.g. instructional dialogues or information querying), we make the assumption that the intrinsic richness of language may be observed and thus has to be taken into account in those systems. As a consequence, possible models for such phenomena should be both computationally valid, since they have to be implemented and integrated within real systems, but also, and foremost, linguistically sound, as they should provide a coverage as good as possible of the variety of cases that real observation may confront us with. In this paper, we will attempt to show how this objective can be realised in the domain of reference resolution, which is a crucial factor of success for a dialogue system with regards its user's acceptance as a good intermediate to achieve their task. More specifically, we try to see to what extent formalising a cognitive model, namely Langacker's Cognitive Grammar (1987, 1991) , is a way to maintain both linguistical soundness and coverage, but also computational validity.

It's a fact that the complexity of reference resolution is due, in part, to the variety of referring expressions, including pronominal reference, definite description, demonstratives, etc. The problem is made even more complex by the apparent variety of mechanisms required to deal with just one of these types of referring expression. For example, the referent of a definite description may be linked to a prior entity within the discourse, "bridged" to a prior entity from which it can be inferred, or accommodated in the discourse domain (Bos *et al.* 1995; Vieira and Poesio 2000).

Moreover, much work on reference centres on pronominal reference (Kamp and Reyle 1993; Lappin and Laess 1994; Grosz *et al.* 1995; Mitkov 1998). As a result, the treatment of other types of referring expressions is typically seen as an extension of or variation on the basic co-referential mechanism involved in pronominal reference. Such an approach, however, does not predict essential differences between the use of pronouns, definite descriptions and demonstratives in contexts where human users would have clear preferences.

Yet, our aim is to design a model of reference resolution to be implemented in human machine dialogue systems. Since it has been shown by psycholinguistic studies (Vivier *et al.* 1997) that it is very difficult and unnatural to impose restrictions on the spontaneous use of referring expressions, we need a unified model of reference, e.g. a model which handles with a single mechanism different types of referring expressions: definite descriptions, demonstratives and pronouns.

One of the backbones of a model for reference resolution is the context model. It is intended to save relevant contextual information for the attribution of referents to referring expressions. Several context models for reference interpretation have already been proposed. Among the best known, Grosz and Sidner (1986) explore the relation between intentional discourse structure and limits of the referential space, modelled as a stack of focus spaces. Centering Theory (Grosz *et al.* 1995) proposes mechanisms for pronominal reference resolution between adjacent discourse units, based on a partially ordered list of entities introduced within them. Discourse Representation Theory (Kamp and Reyle 1993) constructs a global context, comprising all potential referents introduced in the discourse, for which accessibility constraints are defined based on syntactic criteria.

A first problem of these approaches is that linking is considered as the basic operation for referent attribution. As a result, additional mechanisms have to be introduced for other types of relations, such as bridging or accommodation (Lascarides and Asher 1993; Bos *et al.* 1995). However, the systematic preference for linking seems to be questionable not only from a linguistic point of view (Corblin 1987), but also from an empirical one: as shown by Poesio and Vieira (1998), it does not correctly reflect the use of definite descriptions in corpora. Following the authors, about 50% of definite descriptions in a newspaper corpus are used to introduce a new entity in the discourse, and 18% are used as bridging. This means that about 70% of definite descriptions are not actually "linked" to a prior discourse entity. Additionally, we observed through the referential annotation of task-oriented dialogue corpora that it often seems counter-intuitive to link a definite description to a discursive antecedent even mentioned far ahead, when the referent is directly accessible in the visual environment (Salmon-Alt, 2001c). Finally, the linking principle is not entirely suitable for the referential treatment of one-anaphora (*the red one*), other-expressions (*the other triangle*) and ordinals (*the first triangle*). These expressions seem to suppose, rather than a directly accessible entity to which they can be linked, a locally activated context set from which the referent can be extracted.

A second problem concerns the internal structure of the context model. The basic entities of the context models that we have introduced are previously mentioned discourse referents. This seems to be insufficient, because reference resolution relies not only on discursive information, but equally on conceptual knowledge and visual information (Cremers 1996). Furthermore, whereas DRT provides access to all previously mentioned entities, CT considers the previous discourse unit only. However, within the list of identified potential referents, CT provides a precise account of relative salience, whereas DRT specifies only syntactic constraints to narrow the list. Some recent models attempt to apply more precise selection criteria to global discourse (Asher 1993; Hahn and Strube 1997). But they rely on some prior discourse analysis, a strategy that presupposes the ability to automatically recognise discourse structure, and implicitly assumes that discourse analysis precedes reference resolution.

A third problem with these approaches is the context updating operation. We believe that context updating intended to reflect at least some cognitive mechanisms should consist of more than just introducing and linking entities. More precisely, we consider that referring is not only identifying a referent, but also, imposing a particular point of view on the referent and the manner it has been isolated within a set of potential referents. The idea that "a word or an utterance, since it not only specifies the perceived referent but also the set of excluded alternatives, contains more information that the simple perception of the event itself" has already been defended by Olson (1970:265). We assume that this feature should be used to enhance the predictive power of a model of reference calculus. See for instance example (1):

(1) The green block supports the big pyramid but not *the red one*. (Winograd 1972)

Without taking into account the fact of having identified the first block based on its colour, it is indeed impossible to resolve correctly the reference for the elliptic expression *the red one* in example (1). In particular, a heuristic strategy consisting in choosing the nominal head of the most recent noun phrase as an "antecedent" would fail here.

To summarise, the following propositions are prerequisites for any model that attempts to overcome these three problematic points:

– It should take into account all the linguistic variety of referring expressions.

– It should consider that the basic mechanism common to all types of reference is not linking. Rather, it is an extraction from locally activated sets of referents, which are created based on discursive, perceptual and conceptual information.

– It should propose a mechanism of context restructuring which overcomes the standard operations of introducing and linking referents by keeping information about activated context sets and differentiation criteria actually used.

The next section shows how these properties are related to the theoretical foundations of Cognitive Grammar. Section 3 presents the basic principles underlying a cognitive, rather than linguistic model of reference. Section 4 applies the principles to an example.

## 2. Basics of Cognitive Grammar

### 2.1 Some theoretical foundations

Cognitive Grammar (Langacker 1986, 1991) situates linguistic competence within a more general framework of cognitive faculties by assuming that language is neither self-contained, nor describable without reference to cognitive processing. A speaker's linguistic knowledge is characterised as a structured inventory of conventional units: phonological units combine with semantic units to form symbolic units, which may be of any size, from a morpheme to a sequence of sentences (van Hoek 1995).

A first fundamental assumption of Cognitive Grammar is that syntax is not an independent component of linguistic analysis. Basic grammatical categories as well as complex syntactic rules are represented by maximally schematic symbolic units - acquired and adjusted through exposure to actually occurring structures - which are used in constructing and evaluating new expressions. As a second basic assumption, sense is not represented by logical forms. The first reason for this is that semantic structures are characterised relative to knowledge systems that are essentially open-ended. Secondly, the meaning of a given expression cannot be reduced to an objective characterisation of the situation described: equally important for semantics is how the speaker chooses to construe the situation. Therefore, Cognitive Grammar assumes a conceptual rather than truth-conditional semantics, considering that meaning consists of a process of conceptualisation, i.e. activation and restructuring of conceptions in a hearers mind.

More precisely, the conceptualisation of an expression is said to impose a particular image on its domain. A domain is defined as a cognitive structure that is presupposed by the semantic pole of an expression. The particular image imposed by the expression emerges through the profiling of a substructure of the domain, namely that substructure which the expression designates. The profiled subpart of a domain is hypothesised to be more prominent or more highly activated than the domain. However, the semantic value of an expression neither resides in the profile, nor in the domain, but rather in a relationship between the two.

As an example, the semantics of the expression *roof* presupposes the conception of HOUSE and profiles a specific subpart of it (Figure 1a). The expressions *parent* profiles a more abstract conception, being characterised with respect to the conception of a KINSHIP NETWORK (Figure 1b) It is important to note first that any cognitive structure can function as a domain - a concept, a conceptual complex, a perceptual experience, an elaborated knowledge system etc. Secondly, most expressions require more than one domain for their full description. The expression *knife* for example, is characterised, among others, by its shape specification (Figure 1c), its canonical rule in the process of cutting (Figure 1d), its inclusion in a typical place setting with other pieces of cutlery (Figure 1e).

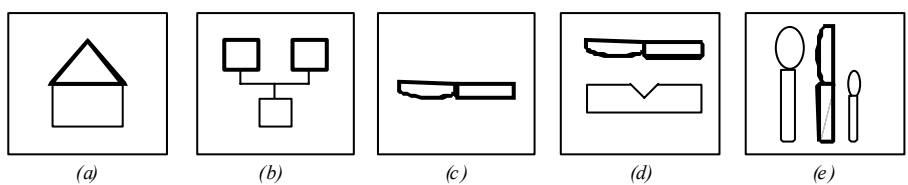

*(a)*     *(b)*     *(c )*     *(d)*     *(e)*

Figure 1 – Profiling domains (Langacker 1991)

## 2.2 Suitability for our purpose

Given our purpose – developing a model of reference resolution suitable for dialogue systems – the biggest problem with Cognitive Grammar is the lack of formalisation. It has, however, several nice properties with regard to the requirements we defined for a cognitive model of interpreting referring expressions (see the end of section 1). We focus here on these properties, before presenting in the next section (3) a model that integrates them into a framework sufficiently formal to have been implemented into a real dialogue system.

The basic assumption about the meaning of linguistic expressions - profiling a substructure within a domain - leads to an explanation of the difficulties induced by considering linking as the basic referential operation. As we suggested before, the fundamental mechanism for interpreting referring expressions seems not to be a linking operation, but rather an extraction from a presupposed domain.

If one accepts this point of view, bridging as well as *one*-anaphora integrate the picture without problems. Indeed, the referent of *the roof* in example (2) is extracted from the conceptual domain introduced by *the house*. In (3), the referent of *the red one* is extracted from a domain of coloured blocks, the same domain from which the referent of *the green block* has been extracted before. Additionally, linking is not excluded definitely, since it can be seen as a particular instantiation of an extraction operation. For example, *the block* in (4) is not considered as directly linked to *the block* mentioned before, but as extracted from a set introduced by *the block* and *the pyramid*.

(2) The house is nice, but *the roof* has to be renovated.

(3) The green block supports the big pyramid but not *the red one*.

Moreover, this manner of considering reference basically as an extraction and not as a linking operation leads to an explanation of the differences between (4) and (5): whereas (4) sounds fine, the repetition of *the block* in (5) seems to be sub-optimal, compared to the use of the pronoun *it* like in (6). The explanation for this observation is the lack, in (5), of a suitable domain from which the referent could be extracted. In this case, the use of a pronoun is preferred. Whereas this phenomenon has been repeatedly noted in linguistic work (Corblin 1987; Gaiffe *et al*. 1997), current implementations do not account for these observations. Furthermore, an explanation based exclusively on "cognitive statuses" of the referents (Ariel 1990; Gundel *et al.* 1993) is insufficient: it is not evident where the difference between the cognitive statuses of *the block* introduced in the first utterances of (4) and (5) is. Consequently, the difference between the uses of a definite description in (4) and (5) cannot be predicted correctly.

(4) The block supports the pyramid. *The block* is big.

(5) The block supports nothing. *The block* is big.

(6) The block supports nothing. *It* is big.

Besides the assumption that linguistic meaning is profiling a sub-structure within a given domain, a second interesting aspect of Cognitive Grammar is the fact that these cognitive domains are not essentially linguistic constructs. Rather, they are based on different knowledge systems, including encyclopaedic and visual information. This property is particularly helpful for dealing not only with bridging references like in the previous example (2), but also with reference to perceptual entities such as in (7), where the referent of *the red block* has not been mentioned before, but is accessible within the visual environment (Cremers 1996). Considering that cognitive domains are built from different knowledge systems allows then to integrate different uses of descriptions (anaphoric, associative, situational – see the classification of Hawkins (1978) into a unified model of reference.

(7) Take *the red block* !

Third, a fundamental claim of cognitive semantics is that the interpretation of an expression is not only the description of a given situation. An equally important fact is construal, e.g. the way that facets of the conceived situation are portrayed. This claim corresponds closely to the one made by Olson (1970:265): "An appropriated utterance indicates which cues or features are critical, while a picture does not. Therefore, there is more information in an utterance than in the perception of an event out of context." Applied to the interpretation of referring expressions, this means that the task is not completely done with the identification of the referent. Rather, it encompasses the identification of a local reference domain and updating the structure of this domain, by profiling the entity designated by the expression. As a result, not only the referent is identified and profiled as the most prominent element of its domain, but also the entire domain is activated, and therefore more accessible for the identification of further referents.

## 3. From Cognitive Grammar to a model of reference resolution

### 3.1 Overview

Following the principles of Cognitive Grammar, we concentrate on a model for reference resolution that attempts to overcome the difficulties discussed in the introduction. The model is based on the fundamental assumption that all reference (independent on the type of the referring expression) is accomplished via access to and restructuring of domains of reference rather than by direct linkage to the entities themselves.

As shown in the previous section (2.2), Cognitive Grammar underlines the need of local context structures. Indeed, an expression is said to be interpreted within a limited domain, presupposed by its semantics, rather than within a global context model containing the list of all previous discourse referents. Therefore, the context representation has to furnish such domains. The next section (3.2) presents a context model built up on domains of reference, which are identifying representations for (possibly partitioned) subsets of contextual entities. We will show in particular that these domains are not primarily linguistic constructs, since they are introduced and updated via discourse, perception or conceptual knowledge.

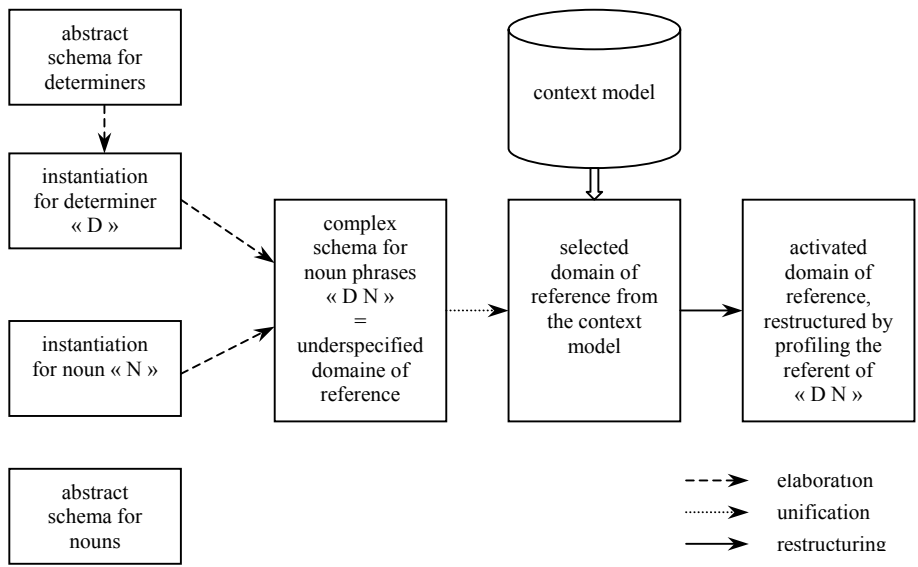

Figure 2 – Overview of the model for processing a referring expression "D N"

Based on a context modelled by domains of reference, we describe in section 3.3 the interpretation process for referring expressions. We adopt here the hypotheses of Cognitive Grammar about the representation of meaning in terms of abstract symbolic schemas: more precisely, we assume that the semantics of a given expression can be represented by a schema which corresponds to an underspecified domain of reference. The underspecified domain itself is calculated by elaborating abstract schemas for nouns and determiners – presupposed in Cognitive Grammar – depending on the semantics of the constituents of the expression being interpreted. The interpretation process properly speaking consists of a unification of the underspecified domain with a suitable domain of reference from the context model and the profiling of a sub-structure – the referent – of this domain (Figure 2).

**3.2 The context model**

*3.2.1 Basic units*

The basic units of our context model are reference domains. Following Sanford and Garrod (1982), Johnson-Laird (1987), Langacker (1991) and Reboul *et al.* (1997), we consider that reference domains are mental representations for entities to which it is possible to refer, including individual objects, collections of objects, events or states. The main difference between a mental representation of an entity and a representation of the entity itself is that a mental representation is not supposed to characterise entirely the entity. This reflects the fact that an expression introducing a new referent does not exhaust the potential features of this referent. Rather, it presents the entity from one or several particular points of view for which we assume in the following that it is the most likely to be activated for referential access to the entity.

Basically, a reference domain is created each time a new entity is introduced in the discourse, but it may also be created for newly perceived entities. The representation of a reference domain consists of attribute-value pairs, minimally including a unique identifier and a type. Type information is derived from a set of generic domains, organised as a type hierarchy, which include general encyclopaedic knowledge or knowledge specific to the application and is assumed to exist prior to discourse processing. Other information provided via the discourse (e.g., specific properties of the object) and via perception (e.g., shape, colour, etc. for objects viewed on a screen) may be added as necessary.

The most important feature of a reference domain are zero, one or more partitions. A partition gives information about possible decompositions of the domain. It could be based on previous discourse information, perceptual information or conceptual knowledge inherited from the generic domains.

The elements of a partition are pointers to other representations, which represent explicitly identified sub-components of the domain. They must be distinguishable from all the other sub-components by the value of a differentiation criterion, which represents a particular point of view on the domain and therefore predicts a particular referential access to its elements. For example, a domain of two marbles (@M, Figure 3) may contain partitions on the basis of colour (a red one and a blue one), position (the one on the right and the one on the left), etc.

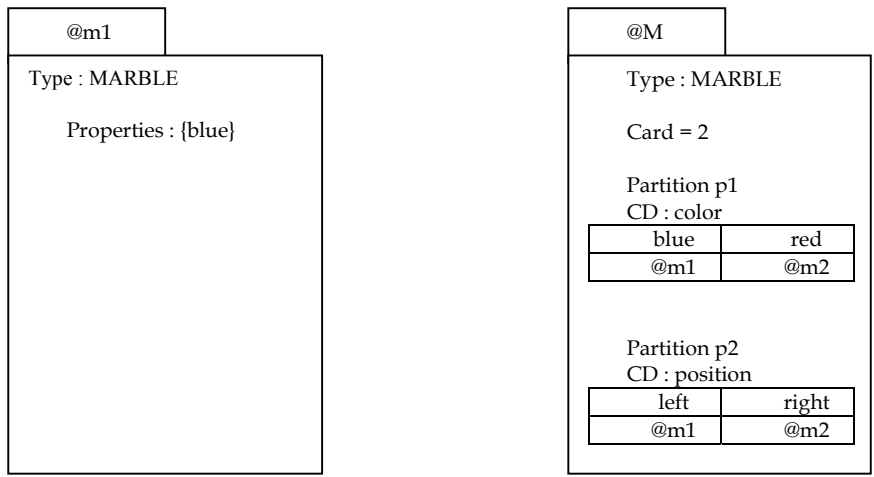

Figure 3 – Reference domain for one blue marble (@m1) and for a group of two marbles (@M)

Within a partition, at most one element may be profiled, according to perceptual or discursive prominence. Profiling is the result of specific operations on the domains, i.e. grouping and extraction. Grouping is briefly discussed below, and extraction, as a part of referential interpretation, is presented in the next section.

*3.2.2 The grouping operation*

The grouping operation is intended to structure the entities of the context model by grouping existing domains into more complex ones. The main goal of this operation is to create new domains and to make them available for the interpretation of referring expressions in the continuation of the discourse. The grouping operation can be triggered by discursive and perceptual factors.

Grouping on discursive factors is defined parallel to the assembly (or elaboration) of complex expressions: in Cognitive Grammar, complex symbols are created by integrating elements at both the semantic and phonological poles. Let us focus on the semantic composition of an example – *the line on the left of the circle*. Figure 4a shows the abstract schema or prototypical meaning of the preposition *on the left of*. It profiles a relationship between two things arranged in a horizontally oriented space. In Cognitive Grammar, relational predications involve an additional prominence asymmetry: the most prominent entity is termed the trajector (tr), and a less prominent one the landmark (lm).

In our example, the abstract schema for *on the left of* is successively elaborated by the representations for *the circle* and *the line*, leading to a composite conception, with the line as the landmark, and therefore the most prominent element (Figure 4b).

Parallel to Cognitive Grammar, the grouping operation of our model maintains the main characteristics of this assembly: it takes two or more domains as the arguments and returns a complex domain with a partition for the grouped elements. Figure 5a diagrams the grouping operation for the same example. Triggered by a discursive factor – here the preposition – the representations for *the line* (@L) and *the circle* (@C) are grouped into a complex domain, partitioned by two properties of the elements: their type and their position. The prominent element of the partition – the representation for *the line* – is the focused element of the domain (indicated by grey background).

Additionally to discursive triggers (preposition, co-ordination, enumeration, arguments of the same predicate), grouping may be triggered by perceptual factors. We consider for instance that perceptual criteria such as similarity or proximity lead to the grouping of contextual entities. Algorithms for grouping visual entities following the principles of the Gestalt-Theory (Wertheimer 1923) can for example be found in Thorisson (1994).

Depending on the type of the grouping trigger, at most one element of a complex domain may be prominent. The treated example – grouping triggered by the preposition on the left of – gives indeed raise to a domain containing a focused entity (Figure 5a). However, grouping does not automatically lead to a focused domain: if the operation is triggered by co-ordination, the resulting domain does not contain any prominent entity (Figure 5b).

To sum up, the context model contains domains of reference, which represent referents or sets of referents. A domain is characterised at least by type information and partitions, providing access to other domains. Partitions are either inherited from generic representations (for example the partition of an entity HOUSE into a ROOF, WINDOWS etc.), or the result of a grouping operation, triggered by discourse or perceptual information. The three fundamental structural characteristics of domains – important to keep in mind for the interpretation process – are the following:

– domain without any partition (@m1, Figure 3) ;

– domain with a partition, but without any prominent item (Figure 5b);

– domain with a partition containing a prominent item (Figure 5a).

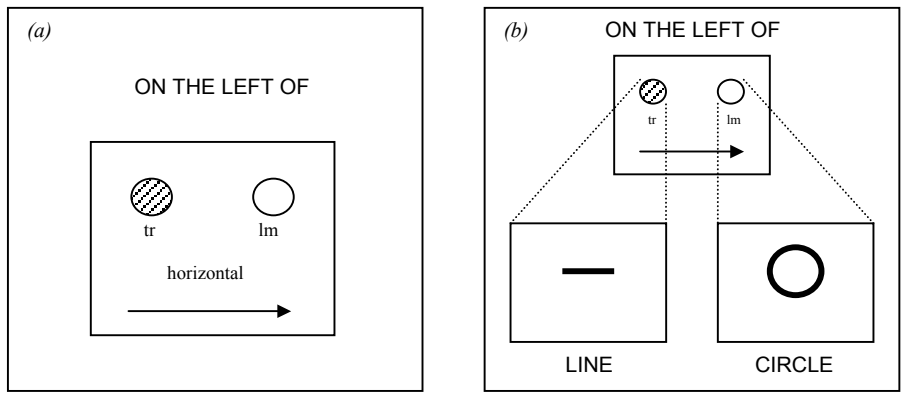

Figure 4 – Assembly of complex expressions in Cognitive Grammar: "the line on the left of the circle"

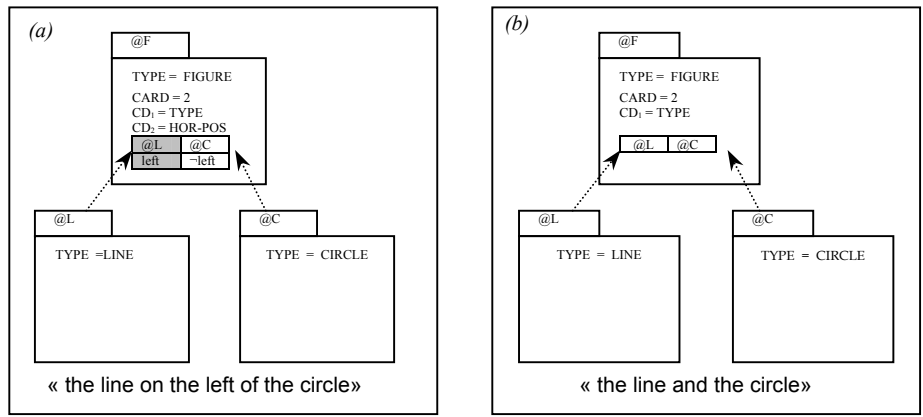

Figure 5 – Grouping operation, triggered by a preposition (a) and by a co-ordination (b)

### 3.3 Interpretation of referring expressions

The context model presented in the previous section provides domains with one of the three fundamental structures mentioned before. Given this context representation, we consider that the role of a referring expression is to select one (or more, in case of ambiguity) of these domains and to restructure it by profiling the referent. The selection operation is constrained by the requirement of compatibility between the selected contextual domain and the underspecified domain construed for the expression being interpreted. In the following sections, we successively present:

– the principles for calculating the underspecified domain depending on the semantics of a given expression (section 3.3.1);

– the selection and unification procedure (section 3.3.2);

– the restructuring operation, leading to the identification of the referent and to an updated and activated domain of reference (section 3.3.3).

The three steps can be considered as going – from left to right – through the diagram of Figure 2.

*3.3.1 Calculus of the underspecified domain of reference*

The underspecified domain associated with a referring expression being interpreted is calculated on the basis of its semantics and depends on two criteria:

– the abstract semantic schema for determiners, elaborated by the semantics of the current determiner;

– the abstract semantic schema for nouns (Figure 6), elaborated by the semantics of the components of the current expression.

The semantics of the determiner associated to the noun then combines with the schema for the noun in order to elaborate a composite representation.

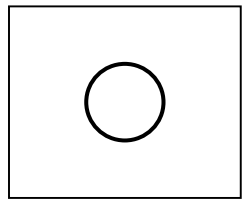     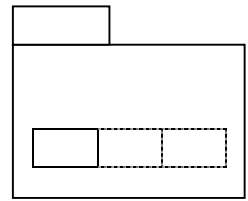

*(a)*                                   *(b)*

Figure 6 – Abstract schema four a noun in Cognitive Grammar (a)
expressed by a partitioned domain (b)

In Cognitive Grammar, determiners are referred to as "grounding predications". They specify the status of nominals in relation to the "ground", comprising the speech event and its participants. Grounding pertains to questions of identification: whether or to what degree the speech-act participants can locate the thing referred to within the mass of objects populating their conceptual universe (Langacker 1991:122 ff. and chapter 12). We adapt this point of view by considering that determiners – precisely because of their power to "locate the thing referred to within the mass of objects populating the conceptual universe" – inform about the composition of the conceptual universe, corresponding in our model to the domain of reference. In this perspective, the determiner, in collaboration with the noun, gives information about the nature and/or properties of the other elements of the partition, i.e. the manner by which the referent (the designated region) is to be isolated within the domain.

Based on these fundamental assumptions, our model accounts for a differentiated treatment of various types of referring expressions within a unified framework, following linguistic principles presented in particular in Corblin (1987) and Kleiber (1994). The next sections comment on Figure 7, showing how the semantics of indefinite, definite, demonstrative and pronominal expressions can be represented by different underspecified domains, which all elaborate the more abstract schema for a noun (Figure 6b).

**Indefinite expressions**

The standard analysis of indefinites (e.g. Russell, or more recently the DRT framework) views them as introducing a new discourse referent, which is independent from both the textual and extra-textual context. In our model, indefinites of the form "a N" (*a circle*) are considered as extracting an element from a domain containing elements of type N. Therefore, they presuppose at the very least such a domain. This can be a specific domain, i.e. a set of Ns given by the context, or per default, the generic (conceptual) domain of N.

However, the idea of context independence is maintained by the fact that they select as their referent arbitrarily an element of the domain. It means that there are no constraints on existing partitions or profiled elements. The dotted structure for "a N" in Figure 7 should be considered as a virtual partition rather than as an existing one, since there is no differentiation criterion, distinguishing the elements N of this partition. It is just intended to indicate that the domain of Ns (specified by the "Type" attribute) must contain at least one element.

To sum up, the underspecified domain of reference for an indefinite noun phrase elaborates the abstract schema for nouns only by adding a type constraint (Type = N). In terms of referential interpretation, this means that an indefinite, in order to be interpreted, presupposes a contextual set (or, per default the generic class) of type compatible elements.

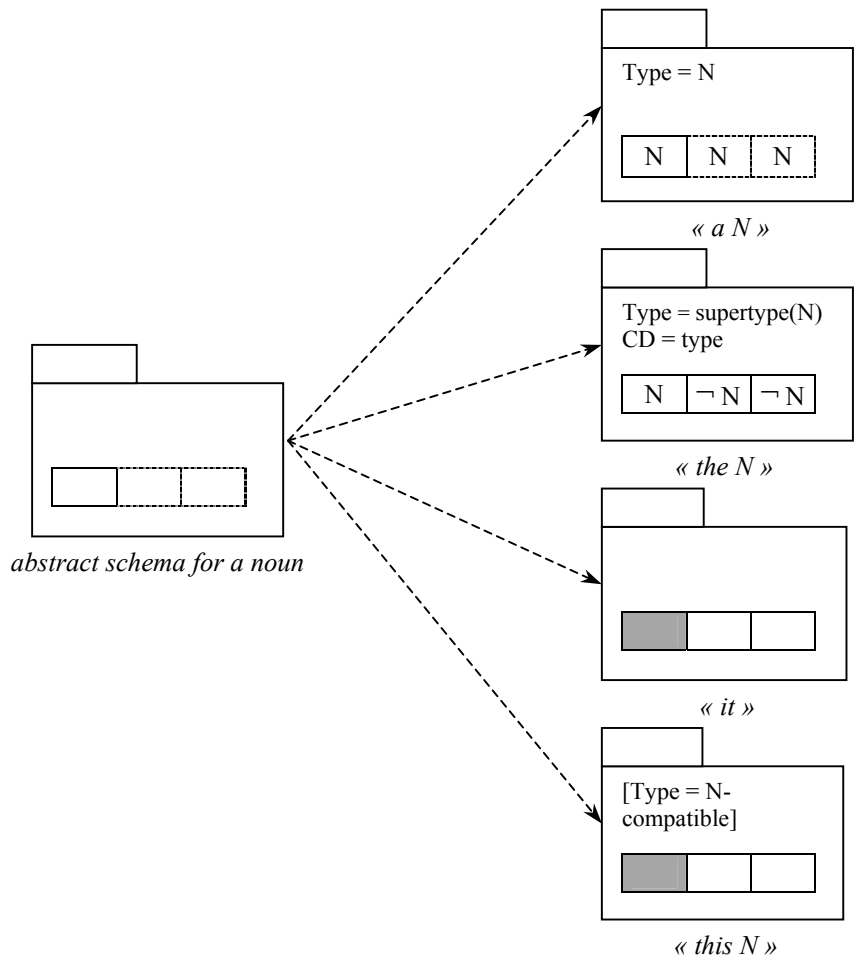

Figure 7 – Building underspecified domains by elaborating the abstract schema for nouns

**Definite expressions**

Current analyses of definite descriptions such as "the N"(cf. for example Bosch and Geurts 1990) assume them to presuppose the existence and unicity of their referent within a discourse model. Our model reflects these intuitions, yet by integrating them into a unified model of reference. Like indefinites and based on the abstract schema for nouns, a definite noun phrase presupposes a domain containing an element of type N. But additionally – in order to guarantee the unicity constraint – the complement of the partition must be distinguishable from that element.

This means that "the N" builds an underspecified domain with, as the most important difference with indefinites, a constraint on an existing partition: the intended referent must be uniquely identifiable via a differentiation criterion. In the case of noun phrases without modifiers, the differentiation criterion (CD) is assumed to be the type designated by "N". As a consequence, the type of the elements of the domain must be a super-type of N, since it comprises elements of type N and of type ¬N.

**Pronominal expressions**

In agreement with most linguistic and computational work on understanding pronouns (for example Kleiber 1990; Grosz *et al.* 1995), we consider that a pronoun refers to an entity being already in the focus. In terms of our model, this means it designates a previous profiled region in some domain. A pronoun being semantically empty, the only constraint on a suitable domain of reference concerns the existence of at least one partition containing a focused entity. In particular, unlike indefinites and definite descriptions, a pronoun doesn't convey any constraint on the type or the differentiation criterion of its domain.

The structure of the underspecified domain is the one sketched in Figure 7: the only elaboration of the abstract schema for nouns consists in adding a focus constraint. This indicates that a pronoun needs a domain containing an already profiled element. Since this element has obligatorily been isolated in previous discourse or via perceptual clues by some differentiation criterion, the partition, unlike for indefinites, is not virtual. However, in comparison to definite descriptions, the nature of this criterion is not significant for the referential interpretation of the pronoun.

**Demonstrative expressions**

Our proposal for treating demonstratives is based on descriptive linguistics for French demonstratives (Corblin 1987; De Mulder 1998). In French, demonstratives seem to differ from demonstratives in other languages, in particular because of their reclassification power. Indeed, a demonstrative expression is said to be able to identify its referent independently of the semantic content of the noun: the semantic content of a demonstrative is therefore available for a reclassification of the referent (Corblin 1987:209). For an example, see (11) onto which we will comment later.

For our model, this means that the only suitable principle for the identification of the referent of demonstratives is focalisation. Unlike indefinites and definites, demonstratives do not convey any constraint on the type of the elements of the domain. Like pronouns, they refer to an entity being already profiled. Therefore, the appropriate domain must contain at least one partition, providing a profiled entity. (Figure 7). The only difference between pronouns and demonstratives is the restructuring operation discussed below (3.3.3).

*3.3.2 Selection procedure*

After the construction of an underspecified domain for the current expression being interpreted, this domain functions as a pattern for the selection of compatible domains from the contextual domains composing the discourse history (cf. 3.2). The selection procedure is modelled by two algorithms. The first one goes through the contextual domains, according to their activation level, starting with the most activated one. The second one is intended to test the compatibility between the underspecified domain and the current contextual domain. Compatibility depends on different criteria: type, cardinality, existing partitions, differentiation criteria and focalisation (Salmon-Alt 2001b). Once a compatible domain has been detected, it has to be unified with the underspecified domain. The resulting domain is returned as the domain of reference. From this domain, the referent will then be extracted by a restructuring operation, described in the next section (3.3.3).

*3.3.3 Restructuring the domain of reference*

Once a suitable domain has been identified among the domains available in the context model, the referent of the expression has to be extracted and profiled. This is done by a restructuring operation, depending on the type of the determiner. Additionally, the result of this operation is different for different s initial domain structures (resumed at the end of section 3.2.2).

Figure 8 gives an overview about the possibilities of restructuring: the first column of the table displays possible structures for underspecified domains, depending on the determiner and introduced in section 3.3.1. The following three column s represent the result of the restructuring operation, each for a different initial structure of the context domain:

– domain without partition;

– domain with partition, but without prominent item;

– domain with a partition containing a prominent item.

In the following, we discuss the restructuring operation more in detail, depending on the type of the expression to interpret.

**Indefinite expressions**

Indefinites are seen as creating systematically a new partition. Within this partition, the referent of the expression is extracted arbitrarily and opposed to the other elements of the domain by a differentiation criterion based on the predicate of the sentence. Furthermore, the extracted referent will be profiled.

The main prediction captured via this operation is the context independence (non-anaphoric use) of indefinites. Indeed, an indefinite expression is compatible with all of the three possible context structures. Additionally, there is no link between the newly created and any existing partition. For example, *a big horizontal line* in A1 of example (8) is extracted from a domain of big horizontal lines where the restructuring operation leads to a partition based on the differentiation criterion "to take" vs. "not to take". In the same way, we predict that a second expression of the same form – see A2'' in (8) – could not be used to refer to the same object. The indefinite in A2'' leads to the creation of a new partition within the domain of big horizontal lines, but since nothing requires to unify these two partitions, the referential identity of the two referents is not guaranteed.

As an interesting point, we predict also that the unification of two partitions created by indefinites must be the linguistically marked case. This is indeed the case of A2''' in example (8). Here, the referential disjunction – and therefore a case of context dependence – for the indefinite *another big horizontal line* is indeed marked explicitly by the modifier *another*.

(8) A1   take a big horizontal line

   B1   yes

   A2   and put *it* on the top of the triangles

   A2'   and put *the big horizontal line* on the top of the triangles

   A2''  and put *a big horizontal line* on the top of the triangles

   A2''' and put *another big horizontal line* on the top of the triangles

**Definite expressions**

For a definite expression *the N*, the item of the suitable type N is identified and profiled within the existing partition of the domain of reference. Note that a definite extracts preferentially a not yet profiled element, so that the preferred restructuring operation for definites seems to be a profile modification.

The interpretation of a definite description is fundamentally context dependant, since it needs a domain with an existing partition in order to isolate an item. This constraint explains why a definite description *the circle* cannot be used for referring to an element of a not yet partitioned domain, such as introduced discursively by *two figures* in A1, example (9):

(9) A1 take *two figures*

    A2 and put *the circle* on the top of *the triangle*

This example is however interpretable, via the presupposition of a suitable partitioned domain, introduced by visual perception. The expression *the circle* can then be seen as selecting the perceptual domain of the figures taken after A1. Within this domain of figures, a partition must distinguish the figures with regard to their type, so that the expression can isolate and profile the referent based on its type.

We also predict the lower acceptability of A2' in example (8), with regard to the pronoun in A2: here, the definite description is interpreted within the domain created in A1, opposing one big horizontal line (profiled) to the others. But since the definite description leads to profiling the already focused element, it does not play the role of a "profile modifier". This correlates with the intuition of an "sub-optimal" use of the definite description, since a pronoun would have the same effect with less effort (Sperber and Wilson, 1986).

**Pronominal expressions**

The restructuring operation for pronouns is empty: a pronoun does not change the structure or the profile of its domain.

Pronouns are therefore seen as indicating the continuity of a domain structure. They need a pre-focused domain and do not change the focus and partition structure. Thus, the typical interpretation principle for pronouns (cf. for example Kleiber 1994; Grosz *et al.* 1995) – referring to objects that are in the focus – is maintained. A2 in example (8) illustrates that. Furthermore, our model does predict awkward cases, for instance A2 of example (10), where the pronominal reference is difficult to interpret, because of the lack of a profiled domain in the context given by A1'.

(10) A1' take a big horizontal line and a small square

    A2 and put *it* on the top of the triangles

**Demonstrative expressions**

*Cognitive Science Quarterly (2000) **1**, ..-..*

Demonstratives take the profiled referent of a given reference domain and change the point of view on this referent by inserting it into a new domain. This feature distinguishes them from pronouns which do not change the structure of the initial domain.

In example (11), the interpretation of *this figure* needs a profiled domain such as introduced by *a big horizontal line* in A1. Within this domain, the expression identifies the profiled element and reclassifies it as an element of type FIGURE by inserting it into a new domain of figures.

(11) A1    take *a big horizontal line*

   B1    yes

   A2    and put *this figure* on the top of the triangles

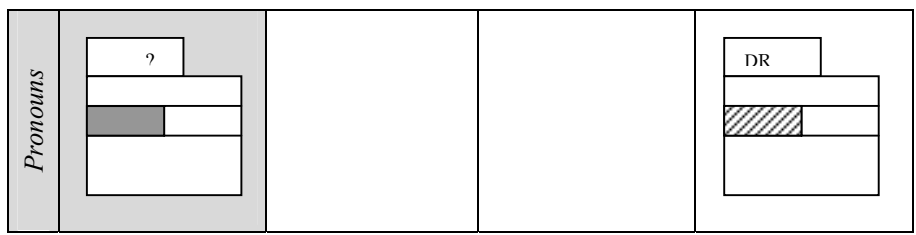

Figure 8 – Compatibility between contextual domains and different types of referring expressions

Furthermore, we predict that demonstratives, like pronouns, are incompatible with non profiled domains. This case is shown in (12) where *this figure* can hardly be interpreted within the domain introduced by *a big horizontal line* and *a small square*.

(12) A1'   take a big horizontal line and a small square

   A2   and put *this figure* on the top of the triangles

## 4. An Example

In order to illustrate the principles introduced in the previous section, we propose now to apply them to the interpretation of the referring expressions of an example, taken from a corpus of task oriented dialogues (Ozkan 1994):

(13) A1   alors tu vas prendre *un gros rond*

   so you have to take *a big circle*

   A2   voilà et à gauche de *ce rond* tu vas prendre *une petite barre*

   okay, and on the left of *this circle*, you have to take *a small line*

   A3   tu ne *la* colles pas *au rond* hein

   don't stick *it* on *the circle*

We start with an empty context. The indefinite *un gros rond* (*a big circle*) leads to the construction of an underspecified domain such as in Figure 9a. Since no suitable domain is available from the context model, the expression is interpreted within the generic domain of BIG CIRCLES (@BC). It extracts its referent from this domain by creating a new partition, characterising uniquely the referent (@bc1) with regard to the other elements of the domain by a differentiation criterion based on predicative information (to take). The resulting context domain is the domain in (b). The underspecified domain for the demonstrative *ce rond* (*this circle*) constraints only the existence of a partition with a profiled element (c). The domain available in the context (@BC) is compatible with this constraint: the profiled element (@bc1) is extracted as the referent and is inserted into a new domain of elements of type *CIRCLE* (d). The indefinite *une petite barre* (*a small line*) is interpreted in the same way as the first indefinite: it profiles an element (@sl1) within the generic domain *SMALL LINE* (f). At this point, a grouping operation intervenes, triggered by the preposition *à gauche de* (*on the left of*). The two contextual elements build up a complex type *FIGURE*, distinguishing their types on, and the small line as the trajector giving bles, cified domain for the pronoun *la* (*it*) imposes a constraint on the existence of a profiled element within the domain (h). @FG provides such a domain: the profiled referent is re-identified, and the domain structure maintained. Finally, the definite *le rond* builds a specified domain such as (j). It constrains the elements type *FIGURE*) and the structure of the which provide of type *CIRCLE*, preferentially not profiled. This domain is compatible with (i), from which @bc1 is extracted and profiled as a referent (k).

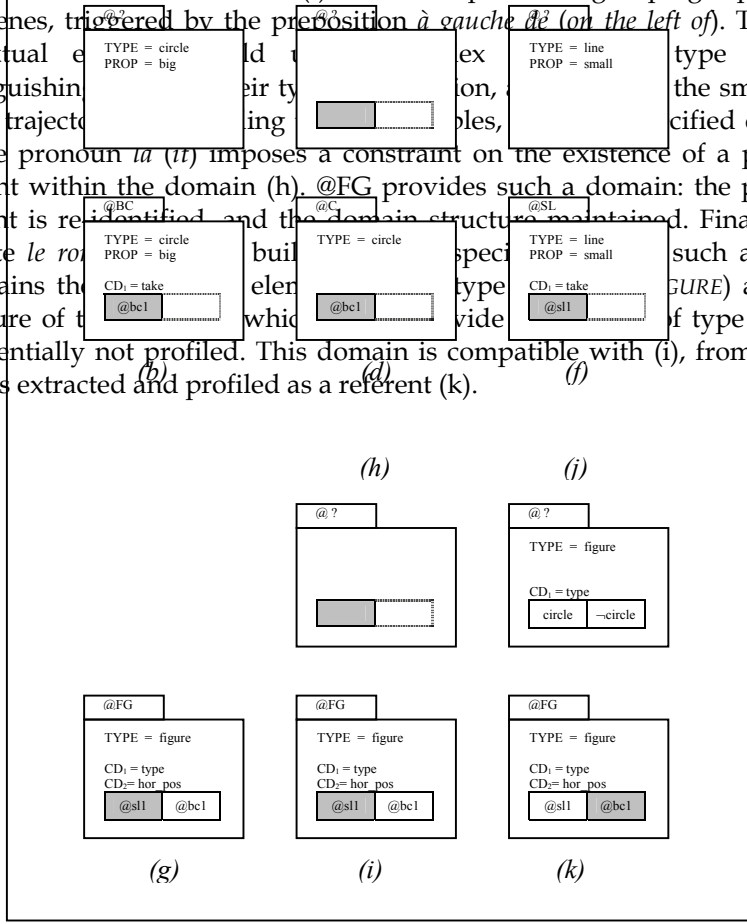

Figure 9 : Interpretation of example (13)

## 5. Discussion

By interpreting a referring expression through a restructuring operation within a domain, we capture the hypothesis of Cognitive Grammar that a referring expression is not only used to identify some entity in a context, but also to convey – through the choice of the determiner and lexical items – a particular point of view on the manner that entity is isolated within the domain. On the base of these principles, synthesised in Figure 8, we are able to make a certain number of predictions, corresponding to the use of referring expressions in natural language production. Moreover, this approach is consistent with linguistic work on referring expressions having shown that each type of referring expression has specific properties, with regard to the presupposed context structures as well as to the interpretative effects (Corblin 1987; Kleiber 1994).

Our model was originally developed to handle reference in a corpus of dialogues concerning the manipulation of figures on a computer screen. Because of the need to deal with all types of reference and different kind of information (linguistic, perceptual and conceptual), our model focuses on cognitive representations and operations rather than primarily linguistic phenomena. However, applying the model to text as well demonstrated that some phenomena that have not been adequately handled in previous work on discourse are easily accommodated in our approach. As the most important feature, we note that our model can with a single mechanism handle what are typically treated as diverse phenomena. Our approach, then, provides at the very least a fresh perspective on the relations between Cognitive Grammar and the problem of reference that may contribute to the overall development of models for reference resolution.